\documentclass{article}

\usepackage{spconf,amsmath,graphicx}
\usepackage{multirow}
\usepackage{tabularx}
\usepackage{array}

\title{Can LLM find the green circle? Investigation and Human-guided tool manipulation for compositional generalization}

\name{Min Zhang$^{\star}$ \qquad Jianfeng He$^{\star}$ \qquad Shuo Lei$^{\star}$ \qquad Murong Yue$^{\dagger}$ \qquad Linhan Wang$^{\star}$ \qquad Chang-Tien Lu$^{\star}$}
  \address{$^{\star}$ Department of Computer Science, Virginia Tech, Falls Church, VA, USA \\
      $^{\dagger}$ Department of Computer Science, George Mason University, Fairfax, VA, USA}

\begin{document}
\maketitle


\begin{abstract}

The meaning of complex phrases in natural language is composed of their individual components. 
The task of compositional generalization evaluates a model's ability to understand new combinations of components.
Previous studies trained smaller, task-specific models, which exhibited poor generalization. 
While large language models (LLMs) exhibit impressive generalization abilities on many tasks through in-context learning (ICL), their potential for compositional generalization remains unexplored. 
In this paper, we first empirically investigate prevailing ICL methods in compositional generalization. 
We find that they struggle with complex compositional questions due to cumulative errors in long reasoning steps and intricate logic required for tool-making. 
Consequently, we propose a human-guided tool manipulation framework (HTM) that generates tools for sub-questions and integrates multiple tools. 
Our method enhances the effectiveness of tool creation and usage with minimal human effort. 
Experiments show that our method achieves state-of-the-art performance on two compositional generalization benchmarks and outperforms existing methods on the most challenging test split by 70\%.

\end{abstract}


\section{Introduction}
\label{submission}

\begin{figure}
    \centering
    \includegraphics[width=3.33in]{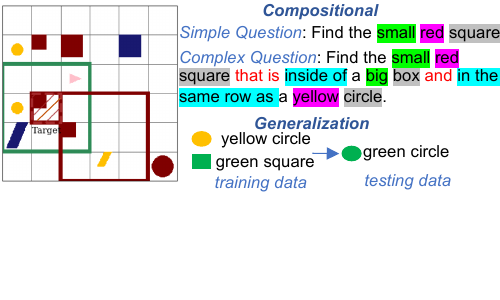}
    \vspace{-20mm}
    \caption{Examples from ReaSCAN dataset. The compositional question is composed of several highlighted individual components. The model needs to generalize to out-of-distribution combinations of previously encountered components.}
    \label{fig:task case}
\end{figure}

Natural languages are compositional, i.e., the meanings of complex phrases are derived from their individual components \cite{partee1984compositionality, harnad1990symbol}. The task of compositional generalization focuses on evaluating a model’s ability to generalize its understanding of components when presented with new combinations.
As shown in Fig.~\ref{fig:task case}, given a grid world and a compositional question, the model is required to output the target object (shaded by inclined lines in the case). The question is composed by several components that are highlighted in the figure. The \textit{simple question} contains no more than 3 components while the \textit{complex question} contains more components and relative clauses. 
The challenge of the task lies in generalizing to Out-Of-Distribution (OOD) combinations that differ from the combinations in the training set. As illustrated in the simple example from Fig.~\ref{fig:task case}, the training data contains "yellow circle" and "green square", while "green circle" in the testing data represents a novel combination of previously encountered components.
In addition to combining simple attributes, the model is also required to combine complicated phrases and relative clauses.
The field has received increased attention from natural language processing (NLP) community \cite{lake2018generalization, kim2020cogs, keysers2019measuring, chen2020compositional, liu2021learning}.

The potential of Large Language Models (LLMs) in compositional generalization remains unexplored. Previous studies trained smaller, specially-designed models, but they cannot generalize effectively on complex OOD questions. LLMs have exhibited amazing generalization abilities on many tasks, via in-context learning (ICL) \cite{qin2023chatgpt, zhao2023survey, kojima2022large}. Therefore, we are enthusiastic about investigating whether LLMs can perform well on the task of compositional generalization.

We conduct an empirical evaluation of prevailing ICL methods in terms of compositional generalization. Under the ICL paradigm, the LLM is given a prompt and directly generates the output for the testing data, without any updates of model parameters. 
The prompt usually contains a few labeled training examples, known as ICL demonstrations. 
We evaluate prevailing ICL methods, including zero-shot ICL \cite{kojima2022large}, standard few-shot ICL \cite{brown2020language}, Chain-of-Thought (CoT) \cite{wei2022chain}, and Program-of-Thought (PoT) \cite{chen2022program}. 


Our investigation \textbf{find} that prevailing ICL methods perform poorly on the task of compositional generalization:
\textbf{(1) CoT can solve simple compositional questions but suffers from cumulative errors in long-step reasoning.} Errors can occur in intermediate steps caused by LLM's hallucinations. \textbf{(2) Although PoT demonstrates a stronger reasoning ability, it struggles to generalize to new syntactic structures within questions.} This limitation arises because PoT attempts to perform all reasoning in a single generation, making it hard to focus on the critical aspects of the question. Despite we want to use external tools to alleviate the reasoning burden, \textbf{(3) creating tools for complex compositional questions is challenging due to the intricate logic}.



To overcome these limitations, we propose a novel Human-guided Tool Manipulation framework (HTM). HTM decomposes a complex problem into multiple sub-problems, analyzes and produces the tools required for each sub-problem, and finally combines these tools to solve the entire question. We incorporate human interactions in the process. 
In this way, the burden of generating one complex tool is reduced to generating and integrating several less complex tools. Using these tools to solve the intermediate reasoning steps not only reduces the cumulative error but also reduces the reasoning pressure in the single generation, i.e., LLMs only need to select the appropriate tools and input the proper argument to the tool. 
The workload for humans is minimal, as they only need to provide feedback or guidance on a small amount of data.

Our contributions are summarized as follows:
(1) We conduct a comprehensive evaluation and point that existing ICL methods perform poorly in compositional generalization.
(2) We propose a human-guided tool manipulation framework, which enhances the success of tool creation and usage with minimal human effort.
(3) Our method achieves state-of-the-art (SOTA) performance on two compositional generalization benchmarks and surpasses existing methods on the hardest test split by 70\%.

\section{Empirical Evaluation of ICL methods}

\textbf{Task Description:}
We focus on ReaSCAN \cite{wu2021reascan} and GSRR \cite{qiu2021systematic} datasets. 
Given a $d\times d$ grid world and a compositional question, the model is required to find the target object (\textit{target identification}) and then generate the action sequence to walk to the target (\textit{navigation}). 
It has been proven that the subtask of navigation can achieve 99.9\% accuracy, given the correct target \cite{sikarwar2022can}. 
Moreover, target identification is more about the compositional generalization of language understanding, but navigation is more about spatial understanding. 
So, in this paper, we focus on target identification and use a predefined module for navigation to generate actions. Our work aims to evaluate and improve the performance of target identification. 
\\
\textbf{Explored Questions:}
We explore a series of issues when applying prevailing ICL methods to the LLM for compositional generalization. Specifically,
(1) Is ICL with LLMs consistently superior to training or fine-tuning smaller task-specific models? 
We compare representative methods from these two categories.
(2) What types of questions have been solved and which types remain unsolved?
We evaluate ICL methods across different OOD testing splits and varying levels of question complexity.
(3) Can LLMs perform effectively on random symbolic expressions? To mitigate potential bias arising from the overlap between testing data and the LLM's pretraining data, we replace original semantic expressions with random symbols rarely present in the LLMs' pretraining data.

Insights and Analysis can be found in section \ref{sec:insights}.

\section{Proposed Method}

\begin{figure}
    \centering
    \includegraphics[width=3.33in]{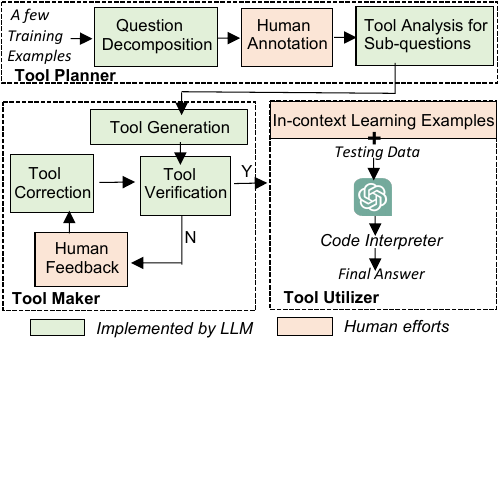}
    \vspace{-30mm}
    \caption{Human-guided tool manipulation framework (HTM). HTM efficiently utilizes LLMs to generate and integrate multiple tools for compositional questions. All human efforts are conducted on a small number of examples.}
    \vspace{-5mm}
    \label{fig:workflow}
\end{figure}

The tool mentioned in our paper represents a Python function that can be executed in a Python interpreter. 
Some existing methods binder human-crafted tools which are labor-intensive \cite{cheng2022binding, gupta2023visual}.
Existing automatic tool-making methods create a single tool for the entire question \cite{cai2023large}. However, the success rate of tool-making is low in our task because complex compositional questions require the tool with intricate logic.
\\
\textbf{Overview:}
To reduce human efforts and simplify tool complexity, we introduce a human-guided tool manipulation framework (HTM), comprising a tool planner, tool maker, and tool utilized. 
As shown in Fig.~\ref{fig:workflow}, the LLM first decomposes a few questions from the training set and analyzes the required tools to solve these sub-questions.
Next, the tool maker utilizes the LLM to generate code for each tool. The tool is fixed under human feedback if it can not pass the validation.
Finally, the LLM learns how to integrate these tools via the in-context learning demonstrations to generate predictions for the testing data.
Remarkably, human efforts through the whole process are minimal, because they only need to provide feedback or guidance on a small amount of data.
\\
\textbf{Tool Planner:} The tool planner is responsible for analyzing the required tools. Instead of creating a single tool for the entire question, a compositional question can be solved by combining multiple simple tools. 
We employ the LLM to decompose the entire question into sub-questions. Human evaluators subsequently assess the feasibility of these sub-questions and provide annotations for their corresponding answers. 
Subsequently, we collect sub-questions related to the same kind of components and let the LLM analyze the required tool based on these sub-questions.
For example, the tool for components highlighted by blue in Fig.~\ref{fig:task case} is $Filter Relationship$. Its input arguments are \textit{head\_objects, condition, tail\_objects} and its output is a list of objects filtered by the condition.
\\
\textbf{Tool Maker:} 
The LLM aims to generate all required tools with the help of human feedback.
For every tool, the LLM first automatically generates the tool according to a few (3 in our experiments) corresponding pairs of sub-questions and labels.
If the generated tool encounters errors or outputs unexpected results when executing in a Python interpreter, the error messages will be appended to the conversation history of the LLM. The LLM will re-generate the tool until it can solve the given examples. 
Subsequently, the generated tool is verified on validation examples (5 in our experiments). 
If the result is not the golden answer, humans will check the reason and give feedback on the wrong logic in the tool. 
Then the LLM will regenerate the tool function with human feedback until the generated tool can pass the verification.
\\
\textbf{Tool Utilizer:}
We teach the LLM how to utilize the generated tool to solve complex questions via ICL.
The prompt contains the description of the tool set, the task instruction, a few-shot ICL demonstrations, and the question of testing instance. 
The description of the tool set includes describing the input arguments, the output, and the purpose of each tool. 
For each demonstration in the prompt, humans write Python code that calls tools to solve the question. The arguments for each tool can be exacted from the question or come from the output of previous tools. We also add code comments to guide the calling of tools.  
For example, we add syntactic parsing in the code comments like \textit{"step 1 objects the small red square has the relative clause led by 'that is', the 'that is' is followed by the step 2 objects"}. In this way, the LLM can understand which referent objects are connected so that proper arguments can be filled for the tool $Filter Relationship$. Following the format of few-shot ICL demonstrations, the LLM can generate code to integrate tools for the testing data.

\begin{table*}
\centering
\caption{Performance on ReaSCAN and GSRR}
\label{table:Semantic results}
\vspace{1mm}
\footnotesize
\begin{tabular}{c|c|*{4}{m{1cm}}|c*{5}{m{1cm}}}
\hline
\multirow{2}{*}{Dataset} & \multirow{2}{*}{Test Split} & \multicolumn{4}{c|}{Training or Finetuning} & \multicolumn{5}{c}{In-context Learning} \\ \cline{3-11} 
                         &     & MM-LSTM \cite{wu2021reascan} & GCN-LSTM \cite{gao2020systematic} & MM-TRF \cite{qiu2021systematic} & Gro-CoT \cite{sikarwar2022can} & Zero-Shot \cite{kojima2022large} & Stand. \cite{brown2020language} & CoT \cite{wei2022chain}  & PoT  \cite{chen2022program} & HTM-(Ours) \\ \hline
\multirow{8}{*}{ReaSCAN} & A1  & 50.4   & 92.3     & 96.7  & 99.6   & 31        & 34.8        & 40   & 96.2 & 98.6      \\
                         & A2  & 14.7   & 42.1     & 58.9  & 93.1   & 24.2      & 25.2        & 30.2 & 93.8 & 97.6      \\
                         & A3  & 50.9   & 87.5     & 93.3  & 98.9   & 28        & 28.6        & 40.2 & 96.2 & 98.6     \\
                         & B1  & 52.2   & 69.7     & 79.8  & 93.9   & 20.2      & 24.6        & 25.4 & 77.8 & 99.4      \\
                         & B2  & 39.4   & 52.8     & 59.3  & 86     & 19.8      & 33.8        & 31.8 & 65.4 & 100       \\
                         & C1  & 49.7   & 57       & 75.9  & 76.3   & 13.6      & 19.6        & 16   & 79.6 & 95.4      \\
                         & C2  & 25.7   & 22.1     & 25.5  & 27.3   & 20.2      & 20.6        & 15.6 & 4.8  & 95.8      \\
                         & AVG & 40.4   & 60.5     & 69.9  & 82.2   & 22.4      & 26.7        & 28.4 & 73.4 & \textbf{97.9}      \\ \hline
\multirow{7}{*}{GSRR}    & S1  & 86.5   & -        & 94.7  & 99.9   & 26.8      & 39          & 37.2 & 91   & 98.8      \\
                         & S2  & 40.1   & -        & 64.4  & 98.6   & 26.8      & 42          & 31.6 & 93.6 & 99.6      \\
                         & S3  & 86.1   & -        & 94.9  & 99.9   & 29        & 33.8        & 38   & 91.2 & 99       \\
                         & S4  & 5.5    & -        & 49.6  & 99.7   & 29        & 39.4        & 35.8 & 89.2 & 98.8      \\
                         & S5  & 81.4   & -        & 59.3  & 99.5   & 24        & 34.4        & 42.8 & 61.8 & 99        \\
                         & S6  & 81.8   & -        & 49.5  & 96.5   & 26.6      & 31.2        & 28.8 & 84.7 & 99.2      \\
                         & AVG & 58.9   & -        & 63.5  & 98.8   & 27        & 36.6        & 35.7 & 85.3 & \textbf{99.1}      \\ \hline
\end{tabular}
\end{table*}

\section{Experiments}

\subsection{Settings}

\textbf{Datasets and Evaluation Metrics:} 
We evaluate the LLM's ability on two compositional generalization benchmarks: ReaSCAN \cite{wu2021reascan} and GSRR \cite{qiu2021systematic}.
The testing set contains OOD combinations not present in the training set. The test splits are shown in Table~\ref{table:Semantic results} and Table~\ref{table:symbolic on ReaSCAN}.
The test splits from A1 to B2 in ReaSCAN and the test splits S1 to S6 in GSRR are mainly about different novel co-occurrences of attributes.
The test splits of C1 and C2 in ReaSCAN involve the generalization of sentence structures, where C1 introduces novel conjunctive clause lengths, and C2 introduces novel relative clauses.
The complexity of questions in test splits P1 to P3 in ReaSCAN increases as the number of relative clauses in the questions increases. 
Each of these test splits comprises 500 test examples in our experiments.
The evaluation metric is the Exact Match (EM) between the predictions and the golden labels.
A higher EM score indicates a stronger ability to generalize compositionally.
\\
\textbf{Setting of Prevailing ICL Methods:}
The in-context learning (ICL) of the LLM doesn't have a training process. We give a few input-label pairs from training examples in the prompt as \textit{ICL demonstrations}. The LLM generates the output for the testing instance directly under a prompt. The grid world is described to the LLM by its attributes and values. 
Zero-shot ICL doesn't have demonstrations while standard few-shot ICL (simplified as Stand.) contains a few demonstrations comprising questions and answers. CoT adds step-by-step textual explanations in each demonstration while PoT generates Python code as the intermediate reasoning steps.
We use 5 demonstrations for standard few-shot ICL. While we use 3 training examples in CoT and PoT, because of the limitation of input length (2048).
\\
\textbf{Models:}
The tool planner and tool maker in our method use GPT-4.
The inference process on testing data for both prevailing ICL methods and our method is conducted on GPT-3.5 version code-davinci-002.

\subsection{Insights of Investigation}
\label{sec:insights}

\textbf{Prevailing ICL methods for LLMs don't always perform better than smaller training or finetuning models.}
We compare the performance of smaller training or finetuning models and prevailing ICL methods for the LLM.
Table~\ref{table:Semantic results} shows the performance over ReaSCAN and GSRR. 
The average EM of the SOTA smaller task-specific networks is 82.2\% in ReaSCAN and 98.8\% in GSRR, while the best performance of the prevailing ICL method of the LLM is only 73.4\% and 85.3\% respectively. 
LLMs need to be well-induced to activate their ability and reach their potential.
\\
\textbf{LLM can handle simple combinations but struggles with complex combinations involving long-step reasoning.}
We test LLM's performance at different question complexities.
As shown in Table~\ref{table:symbolic on ReaSCAN}, CoT achieves 95.8\% EM in P1 while only 19\% in P3, while PoT exhibits consistent performance across P1 to P3.
This is because when looking for objects that meet specific component conditions, like 'red squares', CoT may sometimes produce incorrect results including 'green squares' due to hallucination, resulting in cumulative errors in long-step reasoning. In contrast, PoT executes this process in the Python Interpreter which reduces errors.
However, PoT's performance in C2 drops drastically to only 4.8\% in Table~\ref{table:Semantic results}.
This is because C2 presents novel sentence structures that increase the hardness of reasoning with complex syntactic.
This result suggests that while the LLM can effectively combine simple attributes, it struggles with long-step reasoning questions.
\\
\textbf{The LLM has the ability to compose random symbols.}
We compare the LLM's performance on symbolic expressions and original semantic expressions. Table~\ref{table:symbolic on ReaSCAN} shows that the LLM's performance exhibits a similar trend on both symbolic and original semantic expressions.
The results reveal that the LLM's compositional generalization ability isn't solely based on seeing component combinations during pretraining.

\subsection{Evaluation of Our Method}
\textbf{Main Result:}
As shown in Table~\ref{table:Semantic results}, our method achieves SOTA performance on two datasets. Notably, in the most challenging test split C2 of ReaSCAN, our method achieves impressive EM of 95.8\%, surpassing the prior method GroCoT, which manages only 27.3\%, and the existing ICL methods that attain a mere 20.6\%. 
The limitations of CoT and PoT become evident when confronted with tasks demanding extensive reasoning steps, as they tend to become entangled in excessive details, potentially disturbing the model's parsing of sentence structure. 
In contrast, our method excels in generalization across such challenges.
Within our approach, the utilization of simple tool combinations enables LLMs to focus their attention on higher-level problem-solving, particularly concerning the overall sentence structure. Furthermore, our method incorporates code comments within ICL demonstrations to guide tool utilization and includes syntactic analysis in prompts to assist LLMs in selecting appropriate tool arguments.
Our experimental results substantiate the effectiveness of our approach for compositional generalization. 
\\
\textbf{Ablation Study:} To emphasize the significance of human participation in both tool planning and tool making, we first conduct a comparison of our approach with pure automatic tool generation for the entire question~\cite{cai2023large}. In this evaluation, we generate the toolset five times and achieve a perfect validation pass rate of 5/5 using our framework, while the purely automatic generation approach results in a 0/5 pass rate. This discrepancy highlights the essential role of human intervention in addressing complex questions, as they often require intricate tools with longer code and sophisticated logic.
To further underscore the necessity of human involvement in tool utilization, we evaluate the performance without code comments over the C2 test split in ReaSCAN. In this scenario, the performance dropped to 83.0\%, a reduction of 12.8\%.

\begin{table}[t]
\caption{Comparison of Semantic and Symbolic Performance}
\footnotesize
\label{table:symbolic on ReaSCAN}
\vspace{-3mm}

\begin{center}

\begin{tabular}{l|lcc|ccc}
\hline
- & \multicolumn{3}{|c|}{semantic} & \multicolumn{3}{c}{symbolic} \\
-                  & P1 & P2 & P3 & P1 & P2 & P3     \\
\hline

Zero-Shot              & 78.6 & 28.2 & 20.0 & 67.6 & 17.2 & 14.0    \\
Stand.                 & 78 & 33.6 & 22.0  & 68.8 & 28.6 & 20.4   \\
CoT                  & 95.8 & 43.8 & 19.0  & 97.6 & 37.4 & 21.0   \\
PoT                  & 100 & 98.4 & 97.8  & 94.4 & 88.4 & 81.2   \\
HTM (Ours)                 & 100 & 99.6 & 98.6 & 100 & 99.8 & 98.2    \\

\hline
\end{tabular}

\end{center}
\vspace{-6mm}
\end{table}
\section{Conclusion}
In this work, we investigate the prevailing in-context learning methods in compositional generalization and propose a human-guided tool manipulation framework. Our investigation provides a crucial insight that existing ICL methods struggle to address complex compositional questions effectively. Our innovative approach focuses on planning, creating, and effectively leveraging multiple tools under the guidance of humans.  Remarkably, our method not only simplifies the complexity of generated tools but also elevates overall performance with minimal human intervention. Our experimental results confirm that our method achieves SOTA performance on two compositional generalization benchmarks.

\bibliographystyle{IEEEbib}
\bibliography{_ref/refs}

\end{document}